\documentclass[10pt,twocolumn,letterpaper]{article}

\usepackage{iccv}
\usepackage{times}
\usepackage{epsfig}
\usepackage{graphicx}
\usepackage{amsmath}
\usepackage{amssymb}

\usepackage[pagebackref=true,breaklinks=true,letterpaper=true,colorlinks,bookmarks=false]{hyperref}

\iccvfinalcopy 

\def\iccvPaperID{24} 

\ificcvfinal\fi

\begin{document}

\title{Estimation of Human Condition at Disaster Site Using Aerial Drone Images}

\author{Tomoki Arai$^{1,2}$\hspace{10pt} Kenji Iwata$^1$\hspace{10pt} Kensho Hara$^1$\hspace{10pt} Yutaka Satoh$^{1,2}$\\
$^1$Information Technology and Human Factors,\\
National Institute of Advanced Industrial Science and Technology (AIST) \\
1-1-1 Umezono, Tsukuba, Ibaraki 305-8565 Japan \\
$^2$Graduate School of Science and Technology, University of Tsukuba \\
1-1-1 Tenoudai, Tsukuba, Ibaraki 305-8573 Japan\\
{\tt\small \{t.arai, kenji.iwata, kensho.hara, yu.satou\}@aist.go.jp}
}

\maketitle
\ificcvfinal\thispagestyle{empty}\fi

\begin{abstract}
Drones are being used to assess the situation in various disasters. In this study, we investigate a method to automatically estimate the damage status of people based on their actions in aerial drone images in order to understand disaster sites faster and save labor. We constructed a new dataset of aerial images of human actions in a hypothetical disaster that occurred in an urban area, and classified the human damage status using 3D ResNet. The results showed that the status with characteristic human actions could be classified with a recall rate of more than 80\%, while other statuses with similar human actions could only be classified with a recall rate of about 50\%. In addition, a cloud-based VR presentation application suggested the effectiveness of using drones to understand the disaster site and estimate the human condition.
\end{abstract}

\section{Introduction}
When natural or man-made disasters occur, it is important to quickly 
assess the situation at the disaster site. 
Drones are useful for understanding disaster sites because they can 
quickly search large areas, gather information from a bird's eye view,
and fly over areas too dangerous for humans to enter.
Various studies have been conducted around the world to use drones
to understand disaster sites, including a comparison of flood damage
by Islam  \etal ~\cite{ISLAM2023100225}, assessment of building collapse by
Jiménez-Jiménez  \etal ~\cite{doi:10.1080/19475705.2020.1760360}, 
estimation of wildfire damage maps by Tran \etal ~\cite{rs12244169},
and investigating infrastructure damage from hurricanes by 
Schaefer \etal ~\cite{hurricane01}.
In Japan, aerial images using drones are being taken to understand 
actual disaster sites, and the number is increasing year 
by year ~\cite{drone-jp01}.

In addition to understanding the damage to infrastructure and buildings,
quickly understanding disaster sites is also important to saving lives.
Assessing the condition of victims and rescuers helps determine the order 
of priority for rescue operations and select appropriate rescue methods.
However, observers need to pay close attention to the people they identify 
in aerial images taken from an overhead view, because people may appear smaller 
than buildings and other objects, or their orientation, posture, and clothing 
may change the way they appear.

Therefore, this study investigates an approach to automatically estimate 
human disaster conditions to support rapid understanding of disaster sites
using drones. 
To this end, a dataset of aerial videos of human actions was
created based on a hypothetical disaster in an urban area.

\section{Related work}
\subsection{Use of drone in search and rescue}
Recently, the use of drones for search and rescue of disaster victims 
and people in distress has been explored.
Qi \etal reported on a method that uses a multispectral camera and bio-radar 
to automatically identify the injured and camouflaged fakers ~\cite{app12073457}.
Al-Naji \etal is working on detecting vital signs by detecting chest 
movements of lying survivors from images ~\cite{rs11202441}.
Ono \etal developed fallen persons detection and person shadow detection
method using rotation-invariant features that take into account shadows
and trees in the background region of persons ~\cite{1050577199130429184}.

These studies were conducted in situations where the subjects were unconscious,
collapsed, and immobile. 
When focusing only on rescue, it is highly urgent to
know the location of people who have collapsed and are immobilized, and to
rescue them quickly.
On the other hand, from the perspective of understanding
disaster sites, it is important to understand not only those who have fallen,
but also those who are evacuating to safer locations and those who are calling
for help.

\subsection{Drone aerial video dataset of human actions} 
The situation of the victims is related to their actions. 
If the injuries are minor, the victims can escape on their own. 
However, if the victims are seriously injured and have difficulty walking,
they cannot escape on their own. 
Even if the injury is minor, if the victims are left at the scene,
they can call for help by waving their hands or other means. 
Therefore, it is useful to recognize the actions of the victims in order 
to judging the human condition.

Previous works on drone aerial video datasets of human actions 
include Okutama-Action by Barekatain \etal ~\cite{Barekatain2017OkutamaActionAA} 
and UAV-Human by Li\etal ~\cite{Li_2021_CVPR}.
These datasets primarily record basic human behaviors such as walking and
running, as well as human interactions with each other and with objects, 
and do not consider the background scenario in which the actions occur.
However, the actions of disaster victims can be considered in the context
of the disaster site.
For example, the action of "running" may have different purposes, such as 
"running to save victims" or "running away from a fire.
Therefore, it is difficult to accurately grasp the human condition simply by 
systematically recognizing and classifying only human actions.
Therefore, in order to understand the human condition at a disaster site, 
it is necessary to consider actions in the background scenario of the disaster.

\section{Proposed Dataset}
In this study, a new dataset was constructed by capturing a series of human
actions in a simulated disaster environment with a drone.
The dataset consists of videos of the actions of disaster victims, rescuers, 
passersby, and onlookers in various disaster scenarios.
We captured 62 videos under different disaster conditions, drone altitudes, 
and drone paths.
In 45 of those videos, 20 extras played the actions of people expected in a 
disaster.
All of the extras were dressed appropriately for their roles.

\begin{figure}
\begin{center}
   \includegraphics[width=\linewidth]{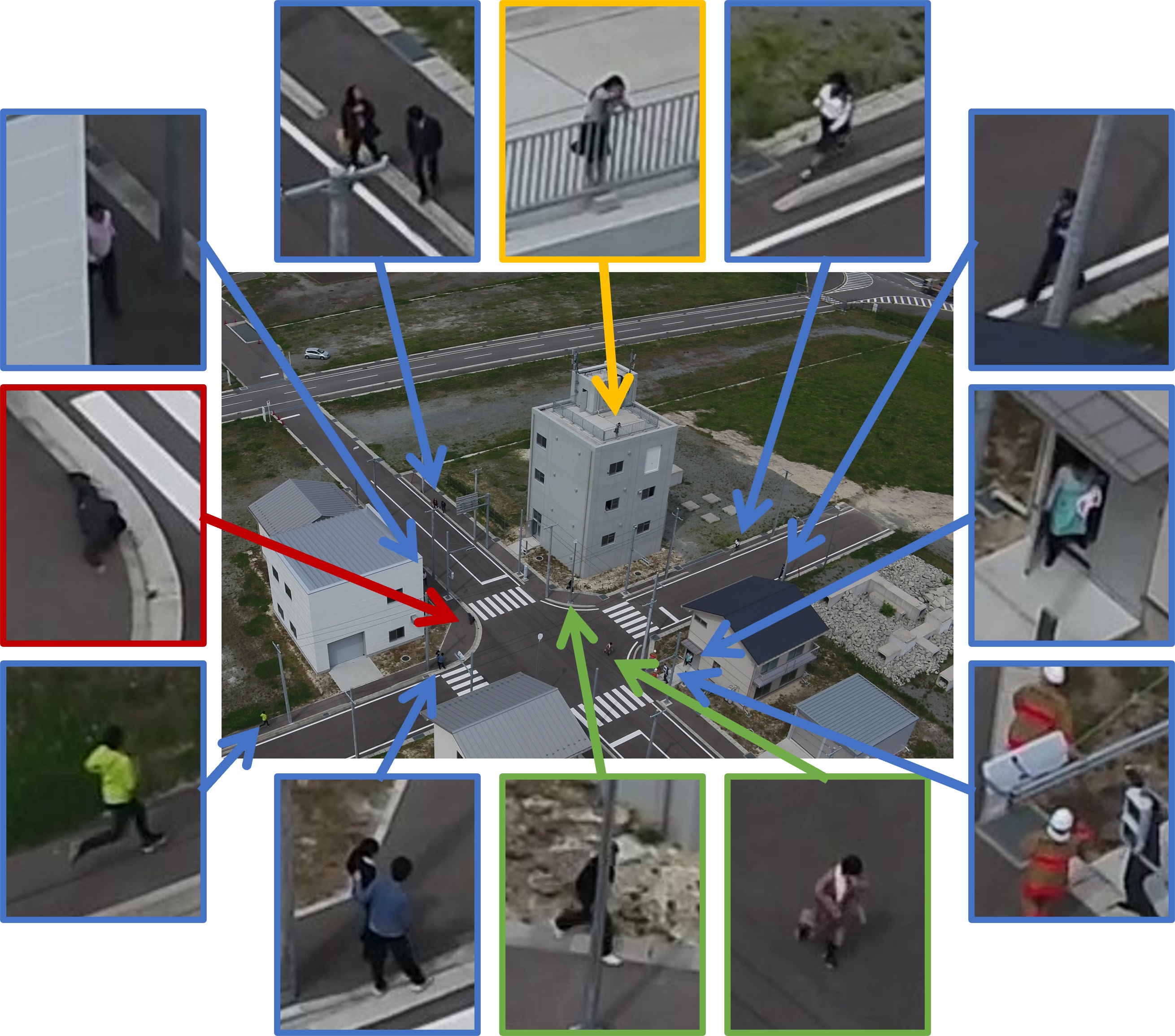}
\end{center}
   \caption{One scene image of the video included in the data set. 
   The drone is turning around the scene at an altitude of 50m and taking
   video. The blue frame shows people in a safe situation, the green frame 
   shows people evacuating, the yellow frame shows people calling for help,
   and the red frame shows people in a dangerous situation.}
\label{fig:video_example}
\end{figure}

The disaster scenario for each video is shown in Table \ref{tab:dataset-ptrn}, 
and the shooting conditions in Table \ref{tab:shooting-cond}.
An example dataset is shown in Figure \ref{fig:video_example}.

\begin{table*}
    \centering
    \begin{tabular}{|c|l|}
        \hline
        Pattern Name & Disaster scenario \\ \hline \hline
        A &\begin{tabular}{l}
                A fire has broken out in a building and people are stranded on the third floor. \vspace{-3pt} \\ 
                On the ground, there are fallen people and onlookers.
                \end{tabular} \\ \hline
        B &\begin{tabular}{l}
                A person is stranded on the roof of a building.  \vspace{-3pt} \\
                On the ground, there are fallen people and onlookers.
                \end{tabular} \\ \hline
        C &\begin{tabular}{l}
                A fire has broken out in a building and many people are evacuated from the building. \vspace{-3pt} \\
                There are people left on the second floor, third floor, and roof of the building, respectively.
                \end{tabular} \\ \hline
        D &\begin{tabular}{l}
                Many people are on the ground or evacuated after the fire. 
                \end{tabular} \\ \hline
        E &\begin{tabular}{l}
                There are people calling for help from different parts of the building.
                \end{tabular} \\ \hline
    \end{tabular}
    \caption{Major assumed disaster scenarios in each video.}
    \label{tab:dataset-ptrn}
\end{table*}

\begin{table}
    \centering
    \begin{tabular}{|l|l|}
        \hline
        Terms & Value \\ \hline \hline
        Location & \begin{tabular}{l}
                  Urban field \\ at Fukushima Robot Test Field
                  \end{tabular} \\ \hline
        Drone altitude &\begin{tabular}{l} 
                        10m, 20m, 30m, 50m 
                        \end{tabular} \\ \hline
        Drone path & \begin{tabular}{l}
                    Straight ahead \\ Turning (small, large radius) 
                    \end{tabular}\\ \hline
        Video Resolution &\begin{tabular}{l} 
                        4K (3840x2160px)
                        \end{tabular} \\  \hline
        Frame rate & \begin{tabular}{l}
                        30fps 
                        \end{tabular} \\ \hline
        Average video duration & \begin{tabular}{l}
                            1min 55sec 
                            \end{tabular} \\ \hline
    \end{tabular}
    \caption{Shooting conditions of the videos in our dataset.}
    \label{tab:shooting-cond}
\end{table}

StrongSORT with OSNet for Yolov5 ~\cite{yolov5-strongsort-osnet-2022}
was used for each video, and bounding box
coordinates and person tracking IDs were automatically generated for persons
in each frame.
Errors in the bounding box coordinates and person IDs were then manually 
corrected using a hand-crafted annotation tool, and a damage status label was
assigned to each person.

\section{Experimental setup}
In this study, four damage statuses, "safe," "evacuation," "call for help," 
and "emergency," were defined as follow in order to clarify the priority order 
of rescue targets, with reference to triage in emergency medical care.

Safe: The person is safe and is not in any of the following three statuses.
Evacuation: The person is evacuated from the dangerous area.
Call for help: The person is calling for help.
Emergency: The person cannot move on his/her own.

These damage statuses strongly influence human actions, but they do not always 
coincide. Therefore, this study attempts to directly estimate human condition 
labels using RGB images containing time-series changes caused by human actions.

In human action recognition, 3D CNNs are frequently used to convolve 2D spatial
and 1D temporal information.
Considering the relationship between human condition and human action, 
this study also uses a 3D CNN for damage status classification.
In this experiment, 3D ResNet-50 by Hara \etal ~\cite{Hara_2018_CVPR} 
was used as the 3D CNN. 

The input clips were generated as follows. 
Each clip consists of 16 frames of one person. Each clip is 
cropped at square coordinates so that the person is in the center of the 9th 
frame. The other 15 frames are cropped at the same position and the annotation 
for the ninth frame is applied in the same way to the other frames in the whole 
clip.

Each of the four classes had 1000 clips.
The number of data for training, validation and testing was 8:1:1.
The number of data for training and validation was randomly selected from Patterns A, C, D, and E,
and the number of data for testing was randomly selected from Pattern B in order to avoid using 
clips that closely resemble the training and test data.

For training, the images were cropped at random positions and resized to 
112x112px. The image was then flipped horizontally at a ratio of 50\%. The 
parameters were set as follows: the loss function was set to cross-entropy loss, 
the optimisation method was set to Momentum SGD (momentum = 0.9, dampening = 0, 
weight decay = 0.001), the learning rate was set to 0.1, the batch size was set 
to 128, and the number of epochs was set to 200. The learning rate was reduced 
to 1/10 for every 50 epochs.
These hyper-parameters were the default values of the 3D ResNet implementation by Hara \etal ~\cite{Hara_2018_CVPR}.

During verification and testing, only resizing to 112×112px was performed.

For comparison with the case of no time series data, experiments were also 
carried out using 2D ResNet. The input image was the 9th frame of each clip 
input to the 3D ResNet and the same parameters were used in the experiment.

\section{Experimental results and discussion}
When assisting in the understanding of a disaster scene, an incorrect assessment 
may result in people at risk going unnoticed. 
Therefore, in this study, the 
recall indicator was used to assess classification performance.

\begin{figure}
  \begin{minipage}[b]{0.45\linewidth}
    \centering
    \includegraphics[width=\linewidth]{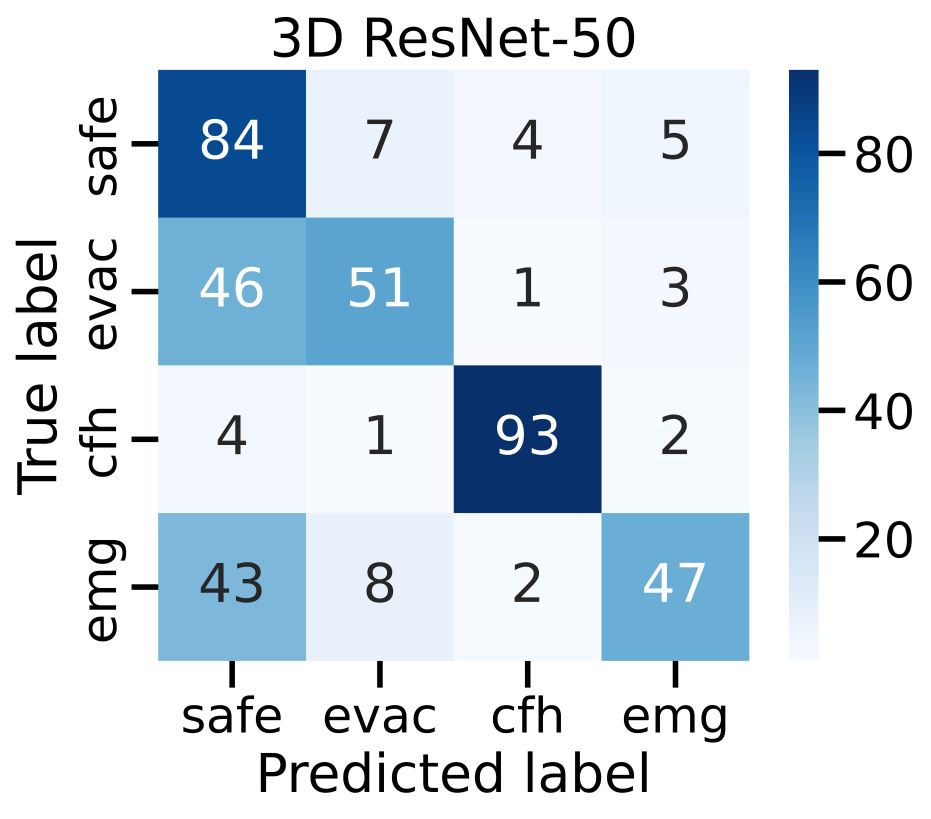}
  \end{minipage}
  \begin{minipage}[b]{0.45\linewidth}
    \centering
    \includegraphics[width=\linewidth]{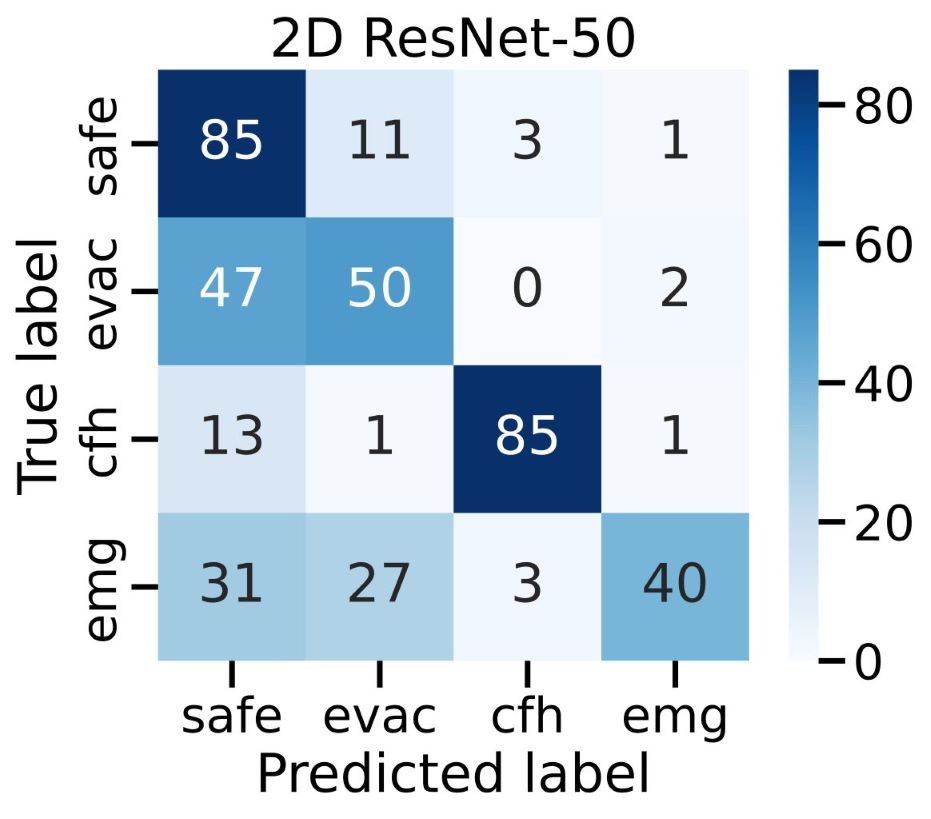}
  \end{minipage}
  \caption{Confusion matrix in 3D/2D ResNet.
  "evac" stands for "evacuation", "cfh" for "call for help", and "emg" for 
"emergency".}
  \label{fig:conf-mat}
\end{figure}

A set of 4000 randomly selected clips was used as one set and the results of the 
three sets were averaged.
The average recall of the test data is shown in Table \ref{tab:result-ds} and 
the confusion matrix is shown in Figure \ref{fig:conf-mat}.

\begin{table}
    \centering
    \begin{tabular}{|l|l|l|}
        \hline
        Class label & 3D [\%] & 2D [\%] \\ \hline \hline
        Safe & 84.33$\pm$3.56 & 84.67$\pm$3.77 \\
        Evacuation & 50.67$\pm$0.41 & 50.00$\pm$0.82 \\
        Call for help & 93.33$\pm$1.47 & 85.33$\pm$2.62\\
        Emergency & 46.67$\pm$4.32 & 40.00$\pm$9.93\\ \hline
    \end{tabular}
    \caption{Recall of human damage status classification in test data.}
    \label{tab:result-ds}
\end{table}

The 3D ResNet classification results for each class showed that "safe" and "call for help" had 
recall values of 80\% or higher, while "evacuation" and "emergency" had recall values of about 50\%.

The reason for this is that "evacuation" and "emergency" are often misclassified as " safe," as shown in the confusion matrix.
The human actions included in "safe" are diverse. 
Although the purposes of the actions are different, the class includes common actions with "evacuation" and "emergency", so it is considered difficult to distinguish them. 
For example, there is the action "running to rescue" in "safe" and the action "running to escape" in "evacuation".
Also "safe" includes action such as being onlookers and not moving, while 
"emergency" includes action such as being injured and unable to move.
On the other hand, "call for help" has unique actions such as waving and is easy to distinguish.

Also, when compared to the 2D ResNet results, the 3D results show that the 
recall of the "safe" and "evacuation" classes remains the same, while the recall 
of the "call for help" and "emergency" classes has improved.
Comparison of the confusion matrices in 3D and 2D showed that there was no 
significant difference in the classification results for "safe" and 
"evacuation", indicating that recall had not changed.
On the other hand, "call for help" and "emergency" with improved recall both 
showed a reduction in misinterpretation of the 3D results, especially 
"emergency", with a significant reduction in misinterpretation to "evacuation".
The reason for this is thought to be that 2D ResNet, which does not learn time-
series data, could not distinguish between the states of moving for evacuation 
and injured and not moving, as both are captured as still images.
In 3D ResNet, which learns time-series data, "emergency" that cannot move on 
their own are misidentified as unmoving onlookers and classified as "safe".

These results suggest that time-series image learning with 3D ResNet strongly 
captures human actions and does not adequately capture features such as the 
purpose of the action, which is sufficient for estimating the human condition.
Therefore, it is important to add features such as location relative to the 
disaster area to achieve higher accuracy in disaster situation classification.
We consider that this would make it possible to distinguish, for example, whether 
the purpose of the action of "running" is to help the victim or to escape 
from the disaster.

\section{Real-time verification}
\begin{figure}
    \begin{center}
        \includegraphics[width=\linewidth]{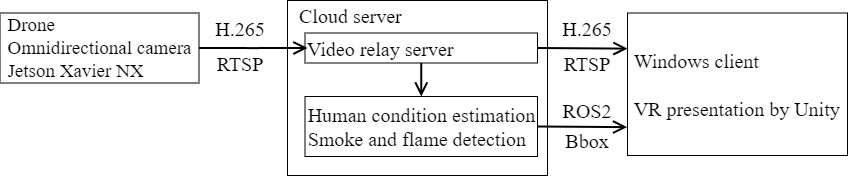}
        \caption{Configuration of real-time damage status classification using 
        the cloud.}
        \label{fig:cloud_config}
    \end{center}
\end{figure}

A demonstration experiment using the damage status classification method 
developed in this study was conducted at the Fukushima Robot Test Field.
We developed a system that transmits real-time images from an actual drone, 
performs image recognition in the cloud, and presents the results on the VR 
space of a client in a remote location.
The configuration of the equipment and software is shown in
Figure \ref{fig:cloud_config}. 
The drone is equipped with a camera (KODAK PIXPR0 4KVR360) capable of 360-degree
omnidirectional 4K video, which is transmitted from the Jetson Xavier NX to a 
video relay server in the cloud via Wifi or 5G communication in 4K H.265 RTSP.
The image recognition program that performs damage status classification and smoke/flame detection runs on the cloud. Images are acquired in real time from a video relay server in the cloud and recognition processing is performed.
The VR presentation program on the client side is implemented using Unity. 
The bounding box coordinates and category ID are acquired from the image 
recognition program via ROS2 communication, superimposed on the omnidirectional 
image obtained from the image relay server, and presented to the remote operator 
as a VR video.

\begin{figure}
    \begin{center}
        \includegraphics[width=\linewidth]{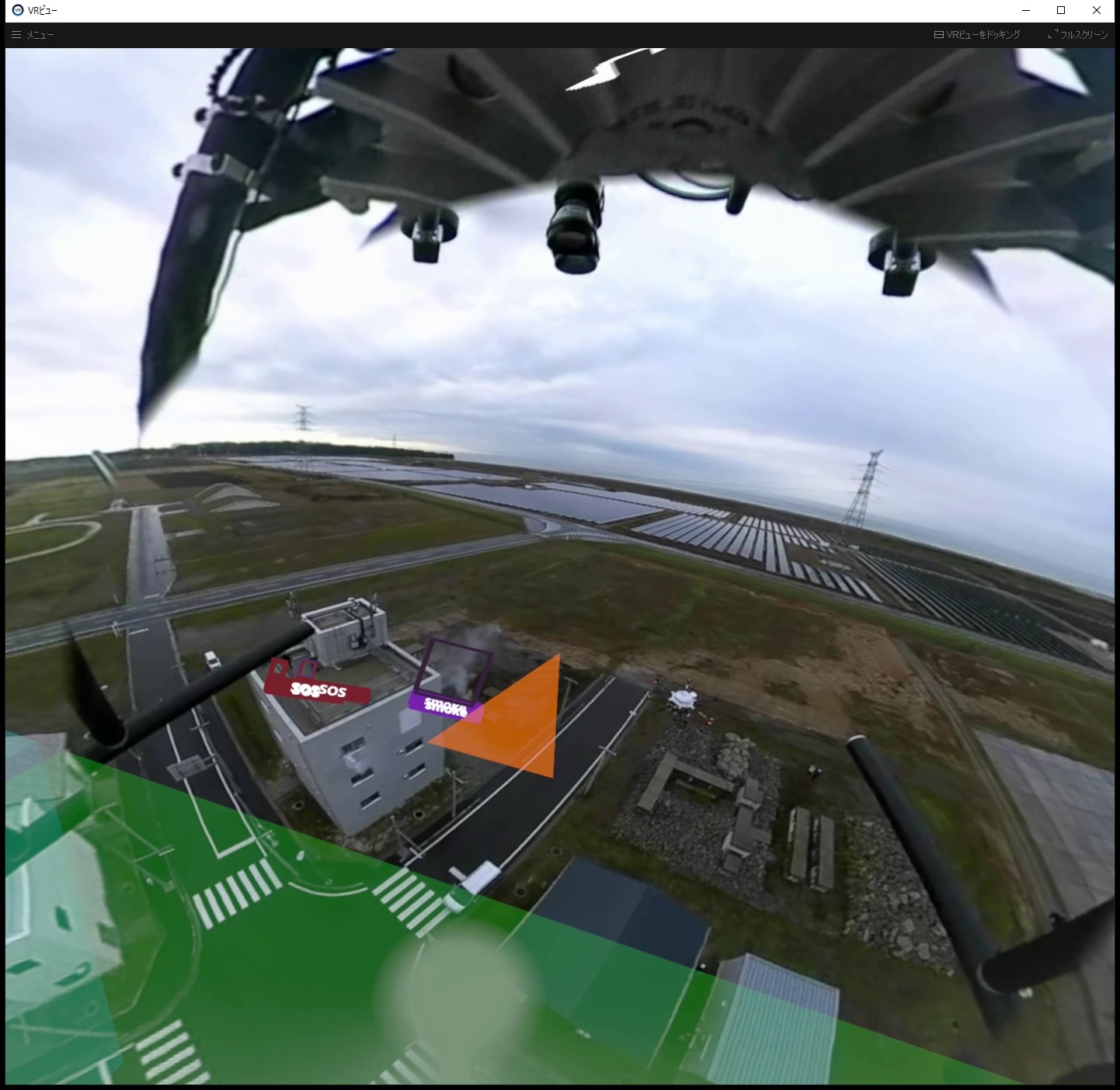}
        \caption{Situation presentation by VR.The results detected as "call for help" 
        are shown as "SOS" in red BBOXes. The detected smoke is highlighted 
        with a purple BBOX.}
        \label{fig:vr_capture}
    \end{center}
\end{figure}

An example of the operation is shown in Figure \ref{fig:vr_capture}.
In this example, smoke is emitted from a window on the third floor of a building 
by a smoke generator, and two people on the roof are waving for help.
Bounding boxes make it easy for the operator to see the results of the waving 
person and smoke detection.

\section{Conclusion and outlook}
In this study, a dataset of human actions in a hypothetical disaster taken 
aerially by a drone was constructed.
By learning time-series images using 3D ResNet, we found that the damage status 
of a person can be classified into "safe" and "call for help" with high recall.
The VR presentation application also suggested the effectiveness of using drones 
to assess the situation at disaster sites.

In the future, we plan to improve the classification by using changes in a 
person's position as an additional feature. A change in a person's position is, 
for example, "evacuate" means moving away from a fire and "safe" means moving 
closer to a fallen person. We believe that by recognizing these changes, the 
system will be able to learn that the same action has different states.
We will also explore changes in performance by searching for hyper-parameters or 
changing the training video pattern and 3D CNN model.

\section*{Acknowledgments}
This paper is based on results obtained from a project, JPNP21004, commissioned by the New Energy and Industrial Technology Development Organization (NEDO).

{\small
\bibliographystyle{ieee_fullname}
\bibliography{reference}
}

\onecolumn
\appendix
\section*{Appendix}
\begin{table*}[h]
    \centering
    \begin{tabular}{|l|l|l|}
        \hline
        Role & Attire & \#people \\ \hline \hline
        Male student & Short-sleeved plain clothes & 1 \\
        Male student & Long-sleeved plain clothes & 1 \\
        Female student & Short-sleeved plain clothes & 1 \\
        Female student & Long-sleeved plain clothes & 1 \\
        Male office worker & Suit & 3 \\
        Female office worker & Suit & 2 \\
        Woman with child & Plain clothes & 1 \\
        Ordinary man & Loungewear & 2 \\
        Ordinary woman & Loungewear & 1 \\ 
        Elderly woman & Long sleeve plain clothes & 1 \\
        Male runner & Runner's style & 1 \\
        Firefighter & Fireproof clothing & 2 \\
        Police officer & Police uniform & 1 \\
        Ordinary man & Long sleeve plain clothes & 1 \\
        Elderly man & Plain clothes & 1 \\ \hline
    \end{tabular}
    \caption{List of extras in the videos of the proposed dataset.
    This table shows the assumed characters and their attire of the extras who 
    played rescuers, victims, etc. in the videos of the proposed dataset.}
    \label{tab:extra-list}
\end{table*}

\begin{figure*}[h]
    \begin{minipage}[b]{0.24\linewidth}
        \centering
        \includegraphics[width=\linewidth]{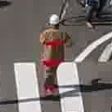}
    \end{minipage}
    \begin{minipage}[b]{0.24\linewidth}
        \centering
        \includegraphics[width=\linewidth]{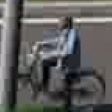}
    \end{minipage}
    \begin{minipage}[b]{0.24\linewidth}
        \centering
        \includegraphics[width=\linewidth]{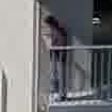}
    \end{minipage}
    \begin{minipage}[b]{0.24\linewidth}
        \centering
        \includegraphics[width=\linewidth]{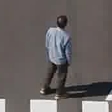}
    \end{minipage} \vspace{10pt}  \\
    \begin{minipage}[b]{0.24\linewidth}
        \centering
        \includegraphics[width=\linewidth]{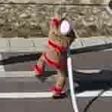}
    \end{minipage}
    \begin{minipage}[b]{0.24\linewidth}
        \centering
        \includegraphics[width=\linewidth]{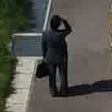}
    \end{minipage}
    \begin{minipage}[b]{0.24\linewidth}
        \centering
        \includegraphics[width=\linewidth]{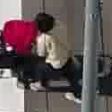}
    \end{minipage}
    \begin{minipage}[b]{0.24\linewidth}
        \centering
        \includegraphics[width=\linewidth]{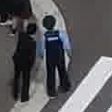}
    \end{minipage}
    \caption{Examples of "safe" class.}
    \label{fig:safe-example}
\end{figure*}

\begin{figure*}
    \begin{minipage}[b]{0.24\linewidth}
        \centering
        \includegraphics[width=\linewidth]{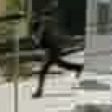}
    \end{minipage}
    \begin{minipage}[b]{0.24\linewidth}
        \centering
        \includegraphics[width=\linewidth]{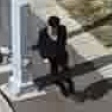}
    \end{minipage}
    \begin{minipage}[b]{0.24\linewidth}
        \centering
        \includegraphics[width=\linewidth]{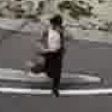}
    \end{minipage}
    \begin{minipage}[b]{0.24\linewidth}
        \centering
        \includegraphics[width=\linewidth]{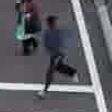}
    \end{minipage} \vspace{10pt}  \\
    \begin{minipage}[b]{0.24\linewidth}
        \centering
        \includegraphics[width=\linewidth]{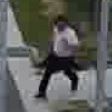}
    \end{minipage}
    \begin{minipage}[b]{0.24\linewidth}
        \centering
        \includegraphics[width=\linewidth]{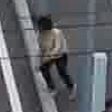}
    \end{minipage}
    \begin{minipage}[b]{0.24\linewidth}
        \centering
        \includegraphics[width=\linewidth]{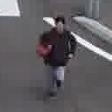}
    \end{minipage}
    \begin{minipage}[b]{0.24\linewidth}
        \centering
        \includegraphics[width=\linewidth]{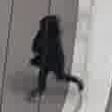}
    \end{minipage}
    \caption{Examples of "evacuation" class.}
    \label{fig:evac-example}
\end{figure*}

\begin{figure*}
    \begin{minipage}[b]{0.24\linewidth}
        \centering
        \includegraphics[width=\linewidth]{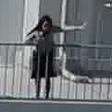}
    \end{minipage}
    \begin{minipage}[b]{0.24\linewidth}
        \centering
        \includegraphics[width=\linewidth]{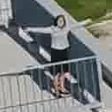}
    \end{minipage}
    \begin{minipage}[b]{0.24\linewidth}
        \centering
        \includegraphics[width=\linewidth]{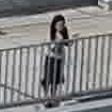}
    \end{minipage}
    \begin{minipage}[b]{0.24\linewidth}
        \centering
        \includegraphics[width=\linewidth]{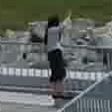}
    \end{minipage} \vspace{10pt}  \\
    \begin{minipage}[b]{0.24\linewidth}
        \centering
        \includegraphics[width=\linewidth]{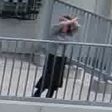}
    \end{minipage}
    \begin{minipage}[b]{0.24\linewidth}
        \centering
        \includegraphics[width=\linewidth]{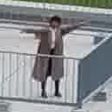}
    \end{minipage}
    \begin{minipage}[b]{0.24\linewidth}
        \centering
        \includegraphics[width=\linewidth]{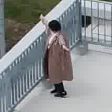}
    \end{minipage}
    \begin{minipage}[b]{0.24\linewidth}
        \centering
        \includegraphics[width=\linewidth]{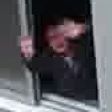}
    \end{minipage}
    \caption{Examples of "call for help" class.}
    \label{fig:cfh-example}
\end{figure*}

\begin{figure*}
    \begin{minipage}[b]{0.24\linewidth}
        \centering
        \includegraphics[width=\linewidth]{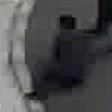}
    \end{minipage}
    \begin{minipage}[b]{0.24\linewidth}
        \centering
        \includegraphics[width=\linewidth]{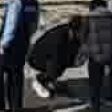}
    \end{minipage}
    \begin{minipage}[b]{0.24\linewidth}
        \centering
        \includegraphics[width=\linewidth]{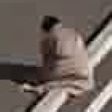}
    \end{minipage}
    \begin{minipage}[b]{0.24\linewidth}
        \centering
        \includegraphics[width=\linewidth]{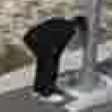}
    \end{minipage} \vspace{10pt} \\ 
    \begin{minipage}[b]{0.24\linewidth}
        \centering
        \includegraphics[width=\linewidth]{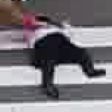}
    \end{minipage}
    \begin{minipage}[b]{0.24\linewidth}
        \centering
        \includegraphics[width=\linewidth]{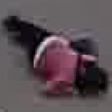}
    \end{minipage}
    \begin{minipage}[b]{0.24\linewidth}
        \centering
        \includegraphics[width=\linewidth]{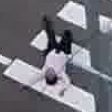}
    \end{minipage}
    \begin{minipage}[b]{0.24\linewidth}
        \centering
        \includegraphics[width=\linewidth]{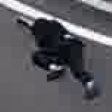}
    \end{minipage}
    \caption{Examples of "emergency" class.}
    \label{fig:emerg-example}
\end{figure*}

\begin{figure*}
    \begin{center}
        \includegraphics[width=\linewidth]{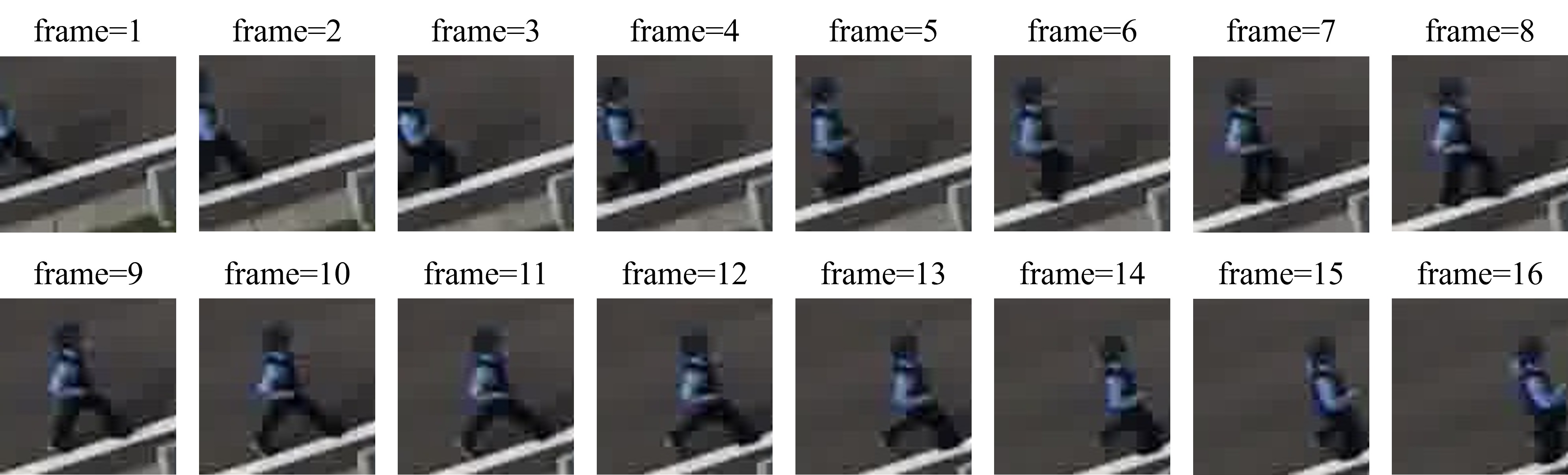}
        \caption{Example of a "safe" class clip.}
        \label{fig:clip_safe}
    \end{center}
\end{figure*}

\begin{figure*}
    \begin{center}
        \includegraphics[width=\linewidth]{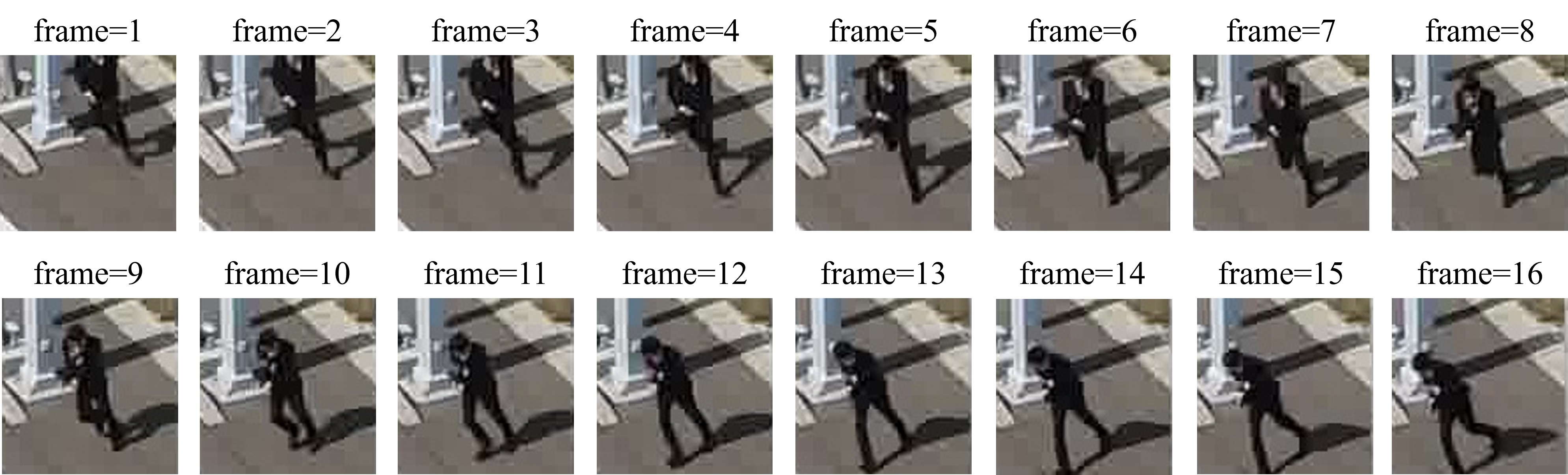}
        \caption{Example of a "evacuation" class clip.}
        \label{fig:clip_evac}
    \end{center}
\end{figure*}

\begin{figure*}
    \begin{center}
        \includegraphics[width=\linewidth]{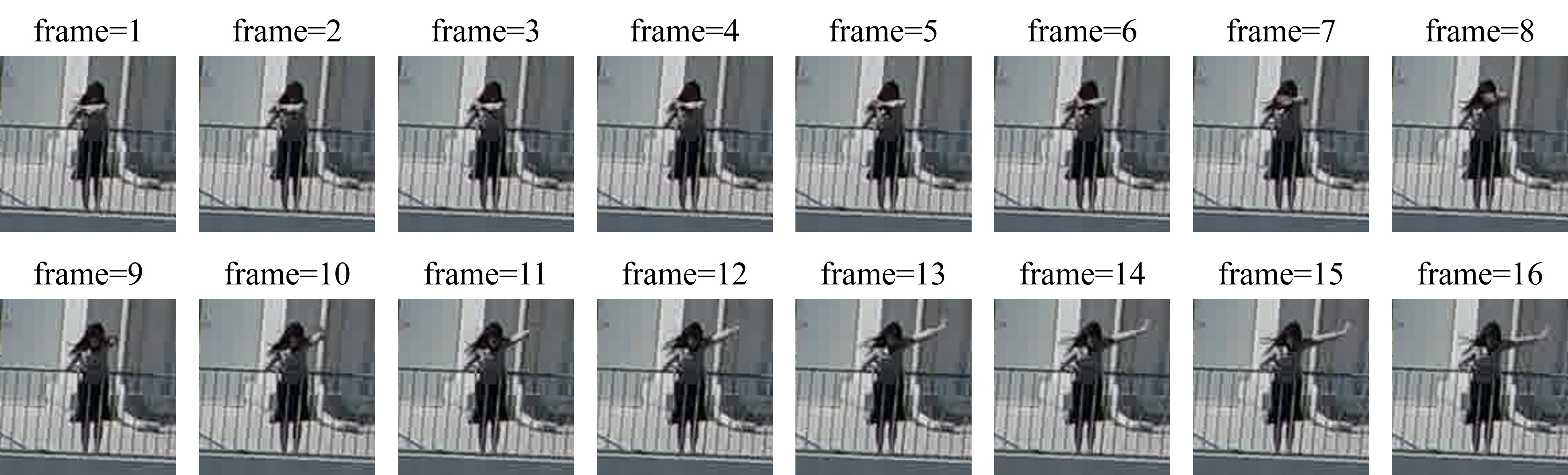}
        \caption{Example of a "call for help" class clip.}
        \label{fig:clip_clf}
    \end{center}
\end{figure*}

\begin{figure*}
    \begin{center}
        \includegraphics[width=\linewidth]{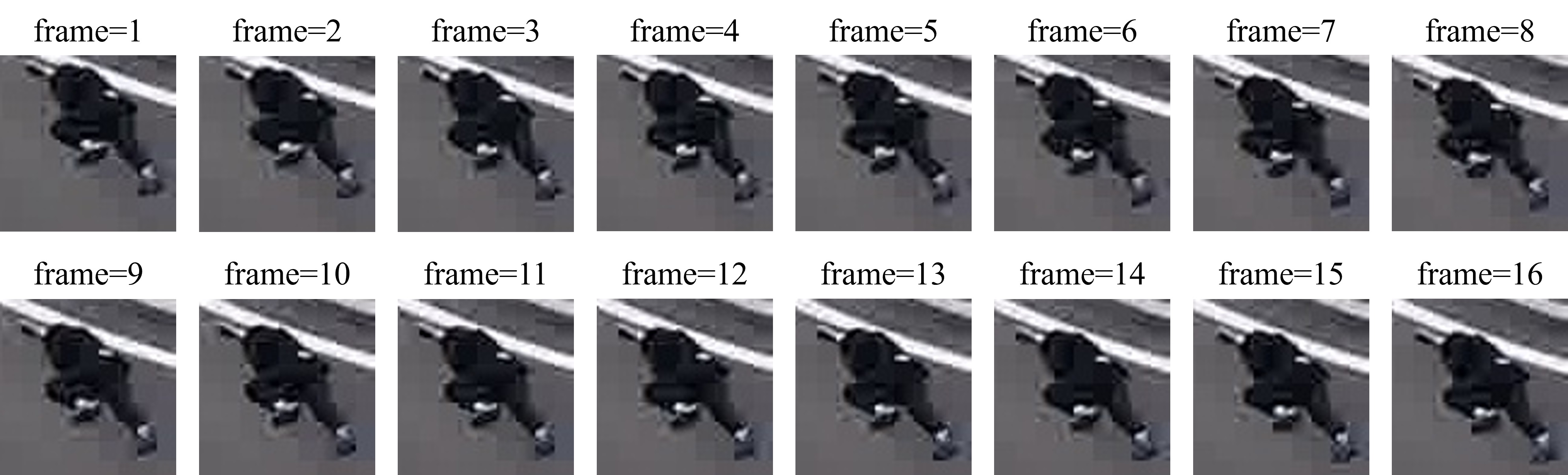}
        \caption{Example of a "emergency" class clip.}
        \label{fig:clip_emerg}
    \end{center}
\end{figure*}
\end{document}